\documentclass[letterpaper, 10 pt, conference]{ieeeconf}  

\IEEEoverridecommandlockouts                              

\usepackage{amsmath,amssymb}
\usepackage[ruled,linesnumbered]{algorithm2e}
\usepackage{algorithmic}
\usepackage{mathrsfs}
\usepackage{color}
\usepackage{cite}
\usepackage{multirow}
\usepackage{url}
\usepackage{xcolor}
\usepackage{xspace}
\usepackage{bold-extra}
\usepackage{subfigure} 
\usepackage{graphicx}
\usepackage{comment}
\usepackage{threeparttable}
\usepackage{arydshln}

\newcommand{\etal}{\textit{et al.}\@}
\makeatletter
\let\NAT@parse\undefined
\makeatother
\usepackage[colorlinks,
            urlcolor=blue,
            linkcolor=blue,       
            anchorcolor=blue,  
            citecolor=green        
            ]{hyperref}

\newcommand{\core}{\texttt{CELL}\xspace}
\newcommand{\sycore}{\mathcal{C}}
\newcommand{\cores}{\texttt{CELL}s\xspace}
\newcommand{\coremap}{\texttt{CELL}map\xspace}

\title{\LARGE \bf
\coremap: Enhancing LiDAR SLAM through Elastic and Lightweight Spherical Map Representation
}

\author{Yifan Duan$^{1}$, Xinran Zhang$^{1}$, Yao Li$^{1}$, Guoliang You$^{1}$, Xiaomeng Chu$^{1}$, \\Jianmin Ji$^{1}$, and Yanyong Zhang$^{2}$*~\IEEEmembership{Fellow,~IEEE}%
\thanks{*The corresponding author.}
\thanks{$^1$ School of Computer Science and Technology, University of Science and Technology of China, Hefei 230026, China {\tt\small \{dyf0202, zxrr, zkdly, glyou, cxmeng\}@mail.ustc.edu.cn, jianmin@ustc.edu.cn}.}
\thanks{$^2$ School of Artificial Intelligence and Data Science, University of Science and Technology of China, Hefei 230026, China {\tt\small yanyongz@ustc.edu.cn}.}%
}

\begin{document}

\maketitle
\thispagestyle{empty}
\pagestyle{empty}

\begin{abstract}
 
SLAM is a fundamental capability of unmanned systems, with LiDAR-based SLAM gaining widespread adoption due to its high precision. Current SLAM systems can achieve centimeter-level accuracy within a short period. However, there are still several challenges when dealing
with large-scale mapping tasks including significant storage requirements and difficulty of reusing the constructed maps. To address this, we first design an elastic and lightweight map representation called \coremap, composed of several \cores, each representing the local map at the corresponding location. Then, we design a general backend including \core-based bidirectional registration module and loop closure detection module to improve global map consistency. 
Our experiments have demonstrated that \coremap can represent the precise geometric structure of large-scale maps of KITTI dataset using only about 60 MB. Additionally, our general backend achieves up to a 26.88\% improvement over various LiDAR odometry methods.

\end{abstract}

\section{INTRODUCTION}

Estimating ego-motion and perceiving the environment are fundamental components of intelligent unmanned systems like unmanned aerial vehicle, unmanned ground vehicle and robots~\cite{cadena2016past}. To achieve this, LiDAR-based odometry and simultaneous localization and mapping (SLAM) have seen significant development.
Specifically, LiDAR-based odometry, aiming at pose estimation, can achieve extremely accurate results over short distances owing to meticulously designed registration methods~\cite{zheng2024traj, vizzo2023kiss} or the effective integration of other sensors~\cite{xu2021fast,xu2022fast}. Comparatively, LiDAR-based SLAM~\cite{pan2021mulls, shan2020lio}, by integrating odometry module and loop detection module, concentrates on constructing a global consistent map.

\begin{figure}[t]
	\centering
	\includegraphics[width = .9\linewidth]{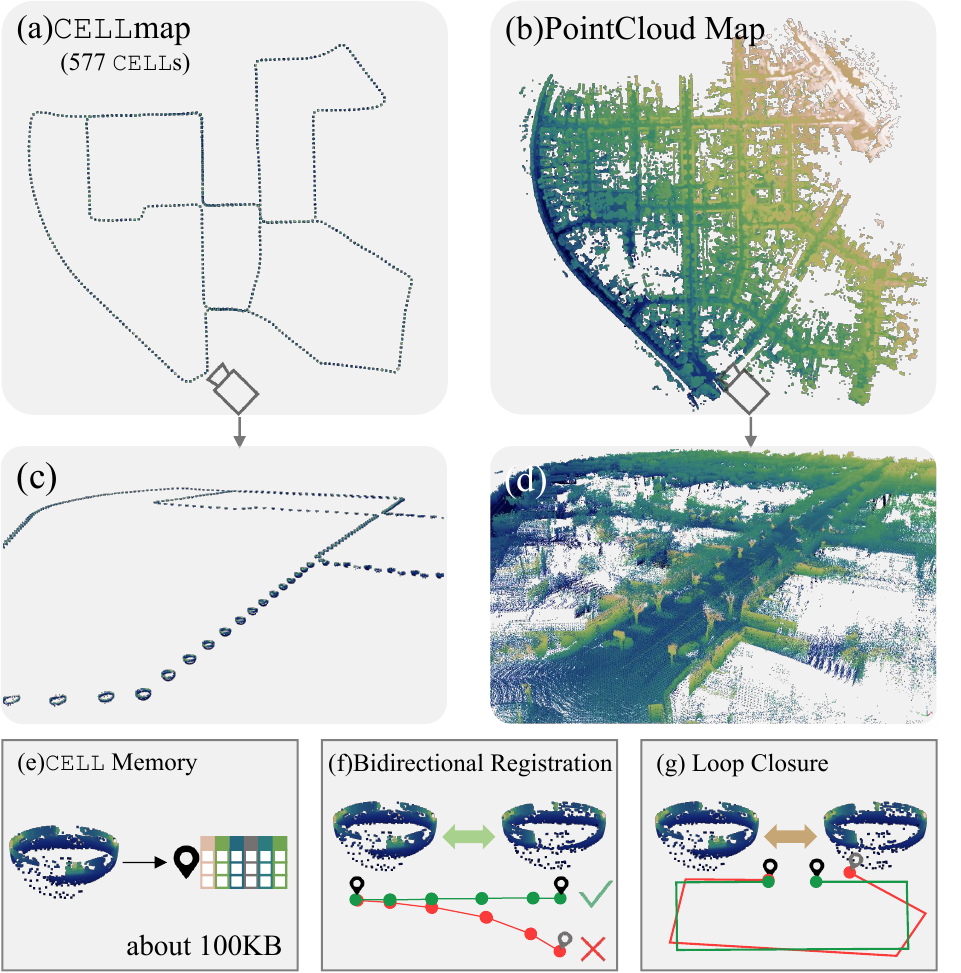}
	\caption{Performance of \coremap on KITTI 00 sequence. (a) The \coremap of KITTI 00  consists of 577 \cores. (b) The corresponding point cloud map. (c) and (d) are zoomed-in views of (a) and (b), respectively. (e) The \core is extremely lightweight, requiring only about 100KB of storage per \core, with the entire KITTI 00 scene needing just 60.2MB. (f) and (g) illustrate a general backend to optimize the \coremap. (f) By introducing more constraints through \core-based bidirectional registration, drift over short distances is reduced, thereby improving pose estimation accuracy. (g) Enhancing global map consistency through loop closure detection. }
	\label{fig:fig1}
    \vspace{-2em}
\end{figure}
However, there are still several challenges when dealing with large-scale mapping tasks. Firstly, using the common map representation like point cloud as an example, As the  size and density of the map grow, the data volume of the point cloud map will rapidly increase, causing inconvenience in both storage and application.
Secondly, to construct a global consistent map, SLAM systems usually need to process the raw data into descriptors~\cite{kim2018scan,meng2024mmplace,fu2024crplace}  for loop detection and retain keyframes~\cite{shan2020lio} or all frames~\cite{liu2023large} to rebuild the map after global pose optimization.
This suggests that a significant amount of data beyond the point cloud map needs to be stored in memory. 
Lastly, due to the disorder of point clouds, they usually need to be managed using KD-Tree~\cite{kdtree} or voxel to support functionalities like registration, which is a crucial and time-consuming step in pose estimation.

Recent methods propose to convert point cloud maps into more compact representations like surfel~\cite{suma, chen2019suma++}, mesh~\cite{vizzo2021poisson, ruan2023slamesh, lin2023immesh} or implicit representation~\cite{pan2024pin}, mitigating the storage growth issue caused by map scale. However, there is still potential for further compression of storage usage. Additionally, because these methods transform point clouds into more compact representations, they inevitably lose some detailed information, resulting in lower trajectory accuracy for most of these methods compared to point cloud-based methods~\cite{zhu2024mesh}.

To address above challenges, we first design an elastic and lightweight map representation called \coremap to replace traditional point cloud maps. As shown in Fig.~\ref{fig:fig1}, the global map \coremap is composed of several \cores, each representing the local map at the corresponding location. 
Specifically, each \core is a spherical data structure, which is why it is referred to as a ``cell.'' The local map is first projected onto a unit sphere with the location of \core as its center. Then, the map point cloud is divided into patches on the spherical surface in a fixed segmentation method to further estimate plane parameters. The geometric structure of the local map can be reproduced using this fixed segmentation method and the extracted plane parameters, therefore we don't need to store the specific locations of each plane like in mesh or surfel. This means that \core is a more lightweight map representation compared to surfel and mesh. Additionally, since \core can represent the geometric structure of local maps, it can also support functions such as registration and loop detection.

During the construction of \coremap, we first utilize open-source LiDAR odometry methods to help build the local point cloud maps and then generate the initial \coremap. 
Next, we propose a \core-based general backend to optimize the \coremap.
Specifically, we design a bidirectional registration method that enhances the accuracy of the applied LiDAR odometry method and a loop closure detection method for global consistent mapping. Since the \core integrates more comprehensive information from the local map than a single scan, while existing LiDAR odometry methods typically use single scan-to-map registration for pose estimation, we implement a \core-based bidirectional registration, i.e., map-to-map registration, to further enhance the accuracy of the LiDAR odometry. Notably, when using \core for registration, we build a unique KD-Tree for all \cores, as they are generated using a fixed segmentation method. This significantly accelerates the registration process, allowing us to use all points of LiDAR scan and perform more iterations.
Additionally, the loop detection module corrects accumulated errors by performing place recognition and pose estimation simultaneously. Based on the optimization results, we adjust only the pose of each \core without modifying the plane parameters of \cores, making it a cost-effective approach for constructing a global consistent map.
 
In our experiments, we use the KITTI dataset as a benchmark to test the proposed \coremap and the general backend. We first employ five common LiDAR odometry methods to generate the initial trajectory for constructing \coremap. By using \core-based bidirectional registration module, the optimized trajectories show up to a 26\% improvement in accuracy. After incorporating the loop detection module, the performance of the SLAM system also surpass that of state-of-the-art LiDAR SLAM methods. For the storage of the \coremap, we only need to save about $0.7\%$ of the original point cloud data to store the global map.

Overall, our contributions are as follows:
\begin{itemize}
 
    \item 
    We propose an extremely lightweight local map representation called \core, which can reproduce the corresponding local map using a fixed segmentation method and plane parameters of the \core without storing bulky point cloud map.
    \item  
    \core integrates more comprehensive local map information than single-frame data, providing more precise constraints in registration.  \core-based bidirectional registration is proposed to enhance the accuracy of various state-of-the-art LiDAR odometry methods.
    \item We implement a loop detection module based on \core, which performs place recognition and pose estimation simultaneously.  Since our global optimization only alters the pose of each \core without changing its contents, this approach incurs minimal cost in achieving a global consistent map.
    \item  To contribute to the community, the code will be open-sourced at https://github.com/yjsx/\coremap.
\end{itemize}

\section{Related Work}
\subsection{LiDAR Odometry and SLAM}
Due to the advancements in LiDAR technology, LiDAR-based SLAM has seen substantial development in recent years. Inspired by LOAM~\cite{zhang2014loam}, numerous variants improve it from different perspectives. For LiDAR odometry, LOAM estimates the pose by extracting edge and surface features from the raw point cloud and performing scan-to-scan registration. Lego-LOAM~\cite{legoloam} and MULLS~\cite{pan2021mulls} further expand the types of features to design different loss functions for various scenarios. Other works~\cite{duan2022pfilter, jiao2021greedy} perform feature selection after feature extraction, removing outliers during the registration process. F-LOAM~\cite{floam} improves the registration method to scan-to-map, enhancing robustness. Additionally, KISS-ICP~\cite{vizzo2023kiss} use direct methods, applying all point clouds without feature extraction, achieving higher accuracy. Other works~\cite{dellenbach2022ct, zheng2024traj} regard LiDAR data as streaming points continuously captured at high frequency and apply continuous-time trajectory estimation. For LiDAR SLAM, an important task is loop detection. To achieve this, descriptors~\cite{kim2018scan, yuan2023std, yuan2024btc} are usually extracted from the raw LiDAR frame to create a descriptor database, enabling the identification of corresponding frames in previously viewed locations.

This paper does not propose a LiDAR odometry algorithm, meaning it does not compute the pose from raw LiDAR data. Instead, it utilizes the output from any LiDAR odometry to construct  \coremap. Additionally, by employing \core-based bidirectional registration and loop detection, we improve trajectory accuracy and ensure the global consistency of the map.
\begin{figure*}[t]
	\centering
	\includegraphics[width = .9\linewidth]{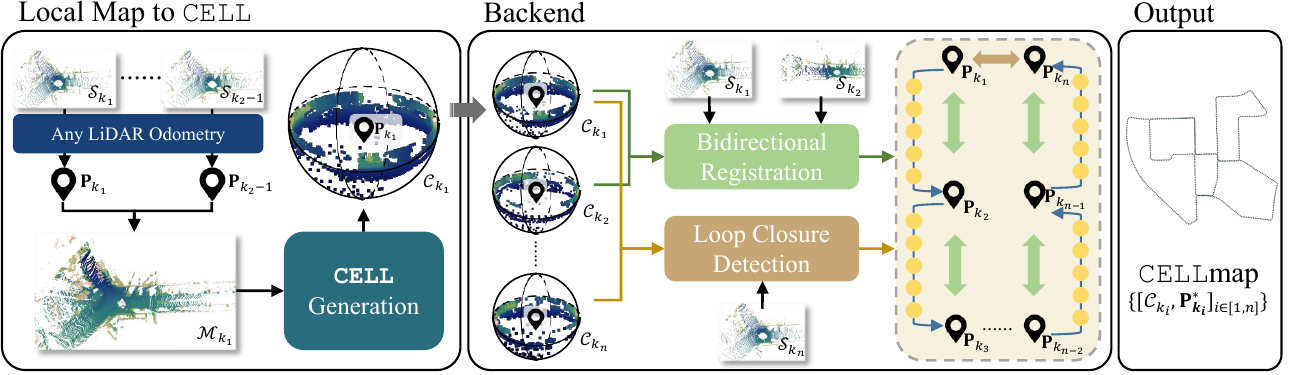}
	\caption{ The overview for constructing \coremap. We first build a local point cloud map using the output from any LiDAR odometry. Then, we generate the corresponding \core through the \core generation module. Next, we enhance the trajectory accuracy of the utilized LiDAR odometry through \core-based bidirectional registration and loop detection algorithms. Finally, we output a globally consistent \coremap.
	}
	\label{fig:pipeline}
    \vspace{-1em}
 
\end{figure*}
\subsection{Map Representation}

In the previous section on LiDAR SLAM, most works use point clouds as the map representation. Some studies explore other more compact representations for map construction. 
Surfel-based map is a more space-efficient representation method compared to point cloud maps. It represents map segments as small disks characterized by attributes such as position, radius, and normal vector. Surfel-based mapping is initially widely used in visual SLAM~\cite{whelan2015elasticfusion, schops2019surfelmeshing}. Subsequently, SUMA~\cite{suma} develops a complete SLAM system that includes odometry and loop detection. SUMA++~\cite{chen2019suma++} further extends it to build semantic maps and filter dynamic objects in the surfel map. Zhang \etal~\cite{zhang2024high} integrate surfel into LiDAR-inertial odometry based on invariant EKF.

Compared to surfel, mesh offers a more continuous map representation composed of points and triangular facets. Vizzo~\etal~\cite{vizzo2021poisson}  represent the map as a triangle mesh computed via poisson surface reconstruction, achieving accurate local maps well-suited for point-to-mesh registration. SLAMESH~\cite{ruan2023slamesh} employs gaussian process reconstruction to expedite the construction, registration, and updating of mesh maps. Mesh-LOAM~\cite{zhu2024mesh} introduces the incremental voxel meshing algorithm, utilizing a parallel spatial-hashing scheme to rapidly reconstruct triangular meshes. However, mesh-based SLAM systems usually do not support loop detection and cannot build globally consistent maps in large-scale environments, as meshes are typically not elastic to loop corrections. 

Additionally, some SLAM methods employ occupancy grids~\cite{occvo,cai2023occupancy}, implicit neural representations~\cite{pan2024pin, zhong2023shine}, and 3D gaussian~\cite{wu2024mm, lang2024gaussian} as map representation. Since the motivations behind these methods differ from those proposed in this paper, they will not be discussed in detail.
The proposed \core is most similar to surfel, as it also preserves the plane information. However, unlike surfel, each plane's position can be recovered through the fixed segmentation method on the sphere, making the \coremap lighter compared to surfel-based and mesh-based maps.
\section{Methodology}

\subsection{Overview}

As shown in Fig.~\ref{fig:pipeline}, the framework of building the \coremap utilizes consecutive frames $\mathcal{S}_i$ from LiDAR as input, while simultaneously retrieving the pose of each frame from any LiDAR odometry methods, denoted as $\mathbf{P}_i$. The frames are stacked according to their respective poses to generate a local point cloud map $\mathcal{M}_i$ and the ending frame is determined by the translation distance between two frames, e.g., $k_1$ and $k_2$ in Fig.~\ref{fig:pipeline}. When the distance between $\mathbf{P}_{k_1}$ and $\mathbf{P}_{k_2}$ surpasses a set threshold, the stacking of the local map $\mathcal{M}_{k_1}$ built from frame $k_1$ to frame $k_2-1$ halts, and the next local map construction begins from frame $k_2$. Then, the completed local map $\mathcal{M}_{k_1}$ is converted into a \core $\sycore_{k_1}$ through a \core generation module, which will be explained in Sec.~\ref{sec：generation}, using the pose $\mathbf{P}_{k_1}$ of the local map as the pose of $\sycore_{k_1}$. The \coremap is defined as the set of \core-pose pairs, i.e., $\{[\sycore_{k_i}, \mathbf{P}_{k_i}]_{i \in [1, n]}\}$.

Due to the inevitable accumulation of errors of any LiDAR odometry, the aforementioned \coremap generated solely from the raw results of LiDAR odometry is imperfect. Thus, we design a general backend to optimize the results of the chosen LiDAR odometry method. 
First, we optimize the poses between adjacent \cores through a bidirectional registration module, introduced in Sec.~\ref{sec：biregistration}. 
Then, a loop closure detection module, detailed in Sec.~\ref{sec：loop_detecton}, is developed that can perform place recognition and pose estimation simultaneously, further enhancing the global consistency of the \coremap. The constraints produced by bidirectional registration and loop closure are then added as factors to the pose graph for optimization. A final \coremap  $\{[\sycore_{k_i}, \mathbf{P}^*_{k_i}]_{i \in [1, n]}\}$ with optimized poses $\mathbf{P}^*_{k_i}$ is the output global map.

\subsection{\core Generation}
\label{sec：generation}
\begin{figure}[t]
	\centering
	\includegraphics[width = .9\linewidth]{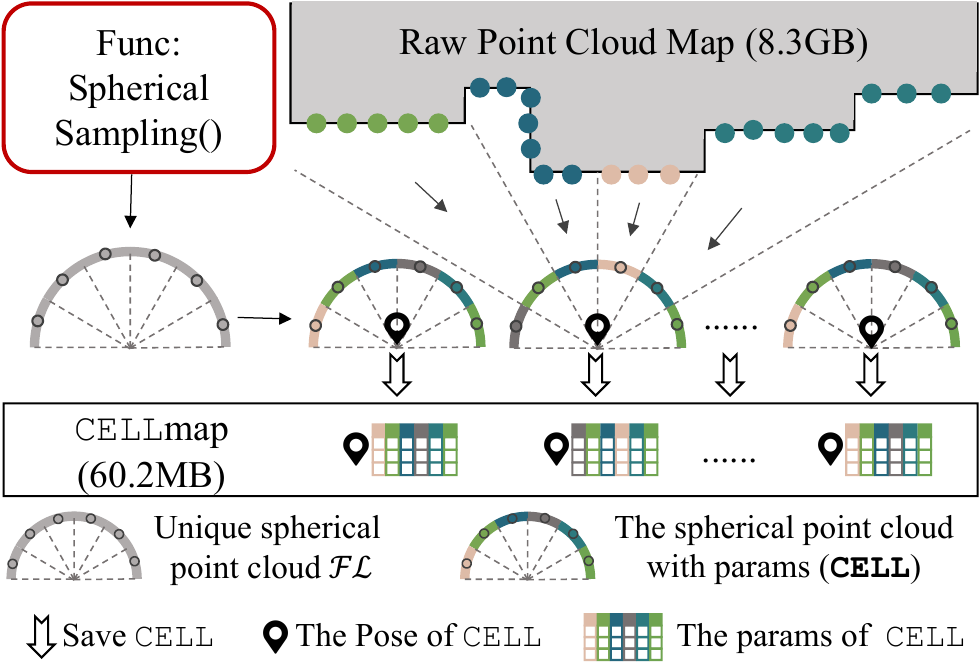}
	\caption{\core generation. We first partition the unit sphere using a unique spherical point cloud $\mathcal{FL}$. Then, each time a \core needs to be generated, the corresponding local point cloud map is segmented according to $\mathcal{FL}$, and the plane parameters for each segment are extracted. Finally, each \core is composed of an $N_{sp}*4$ array.
	}
	\label{fig:generation}
    \vspace{-2em}
 
\end{figure}
For point cloud maps, the discreteness of point cloud necessitates a high density to preserve map details, resulting in significant storage and computational costs. Some methods use spatially continuous and more compact representations like mesh or surfel instead of point clouds to reconstruct the surfaces. However, we wonder if further compression is possible.
Using surfel as an example, each surfel is defined by a position $\mathbf{v} \in \mathbb{R}^3$, a normal $\mathbf{n}\in \mathbb{R}^3$ and a radius $r \in \mathbb{R}$~\cite{suma}. 
If the position information can be omitted, the storage required for surfels will be nearly halved.

\begin{figure}[t]  
    \centering   
    \begin{minipage}[t]{0.48\linewidth}  
        \centering  
        \subfigure[Local point cloud map, colored by z-value] {\includegraphics[height=2.5cm]{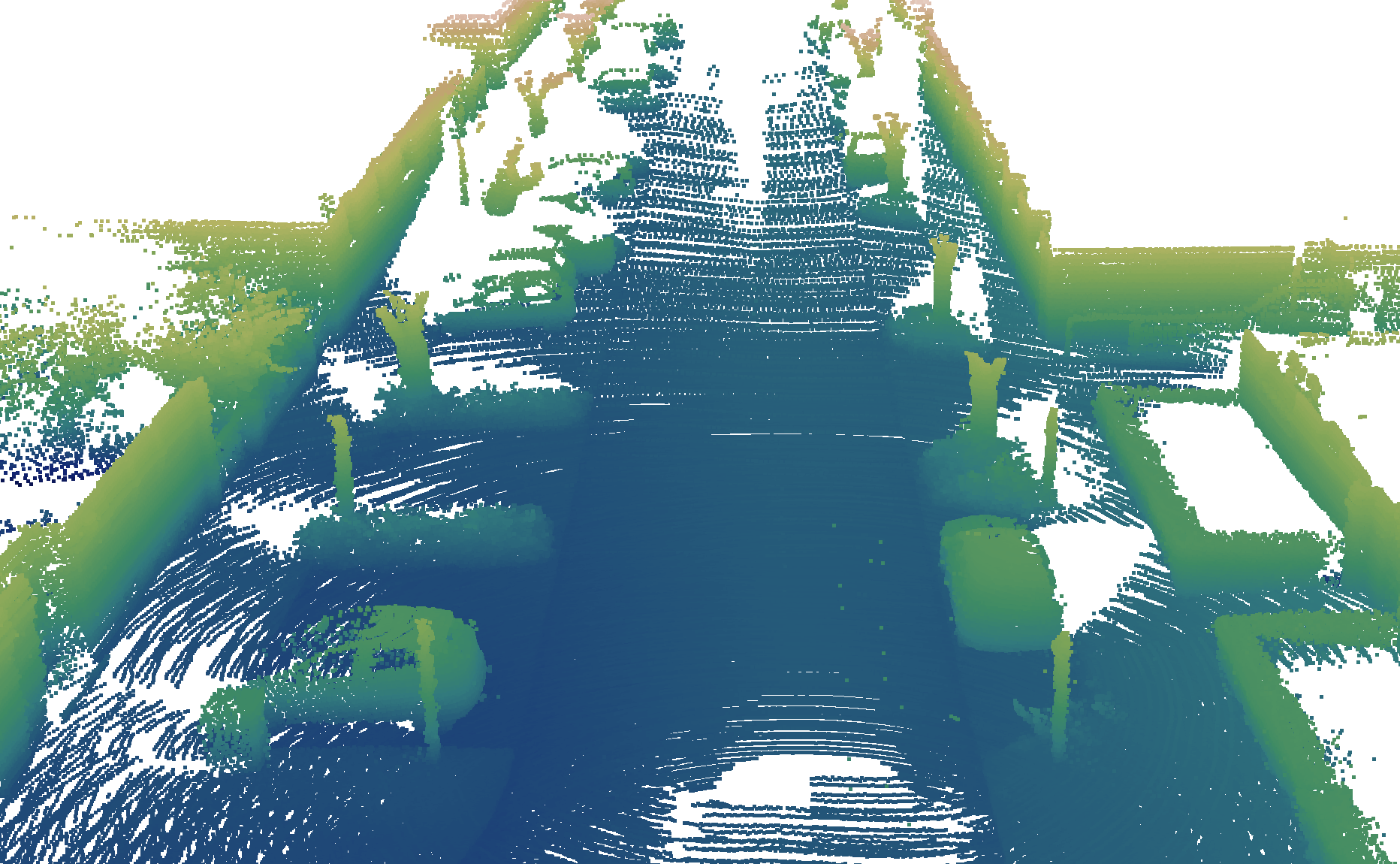}} 
    \end{minipage}  
    \hfill   
    \begin{minipage}[t]{0.48\linewidth}  
        \centering  
        \subfigure[The unique spherical point cloud $\mathcal{FL}$] {\includegraphics[height=2.5cm]{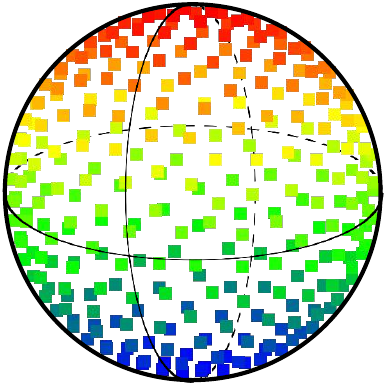}}  
    \end{minipage}  
    \vspace{0.5cm}    
    \begin{minipage}[t]{0.48\linewidth}  
        \centering  
        \subfigure[Segmentation on the local map] {\includegraphics[height=2.5cm]{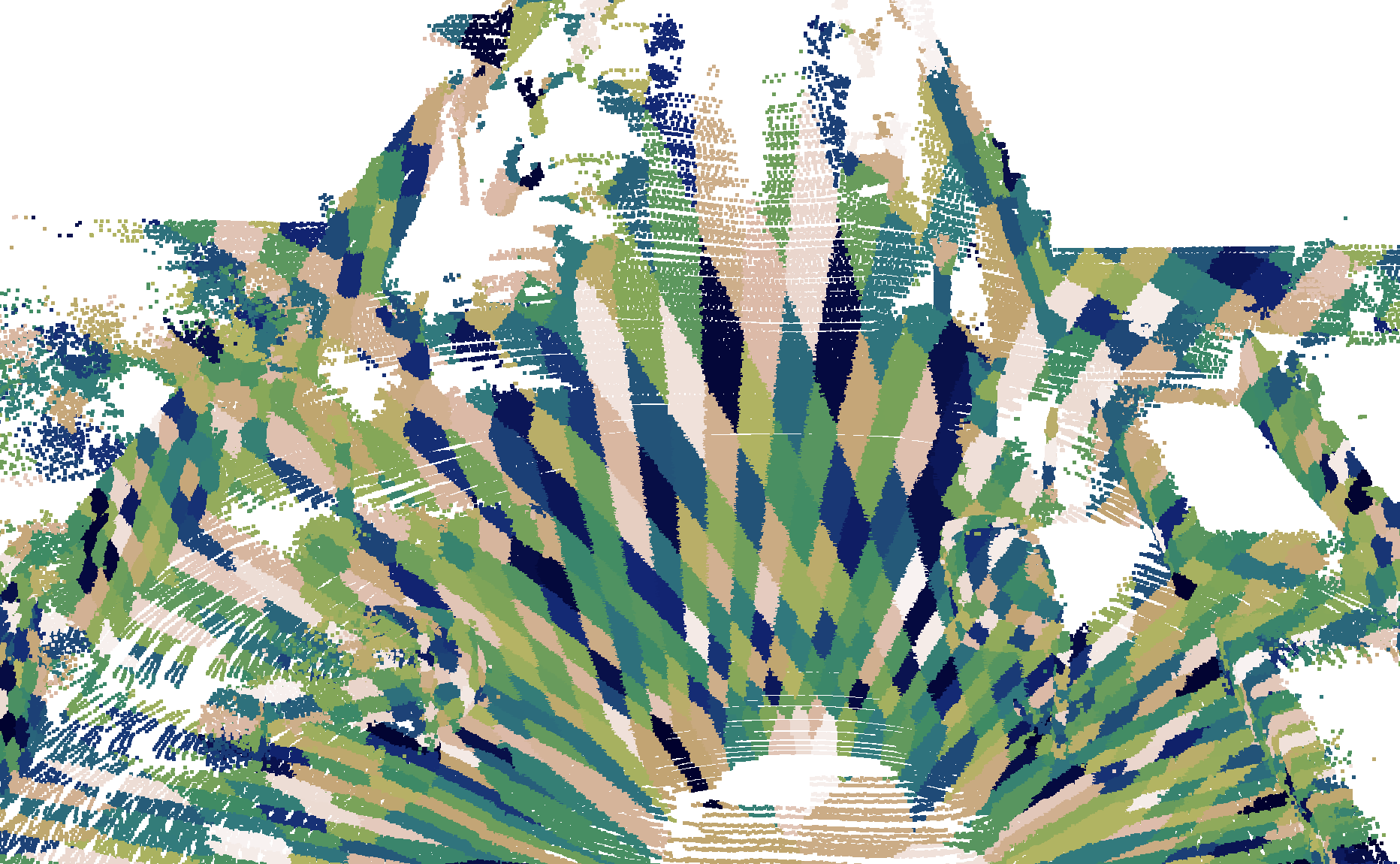} \label{fig:segment}} 
    \end{minipage}  
    \hfill   
    \begin{minipage}[t]{0.48\linewidth}  
        \centering  
        \subfigure[Normal of each point according to the extracted plane] {\includegraphics[height=2.5cm]{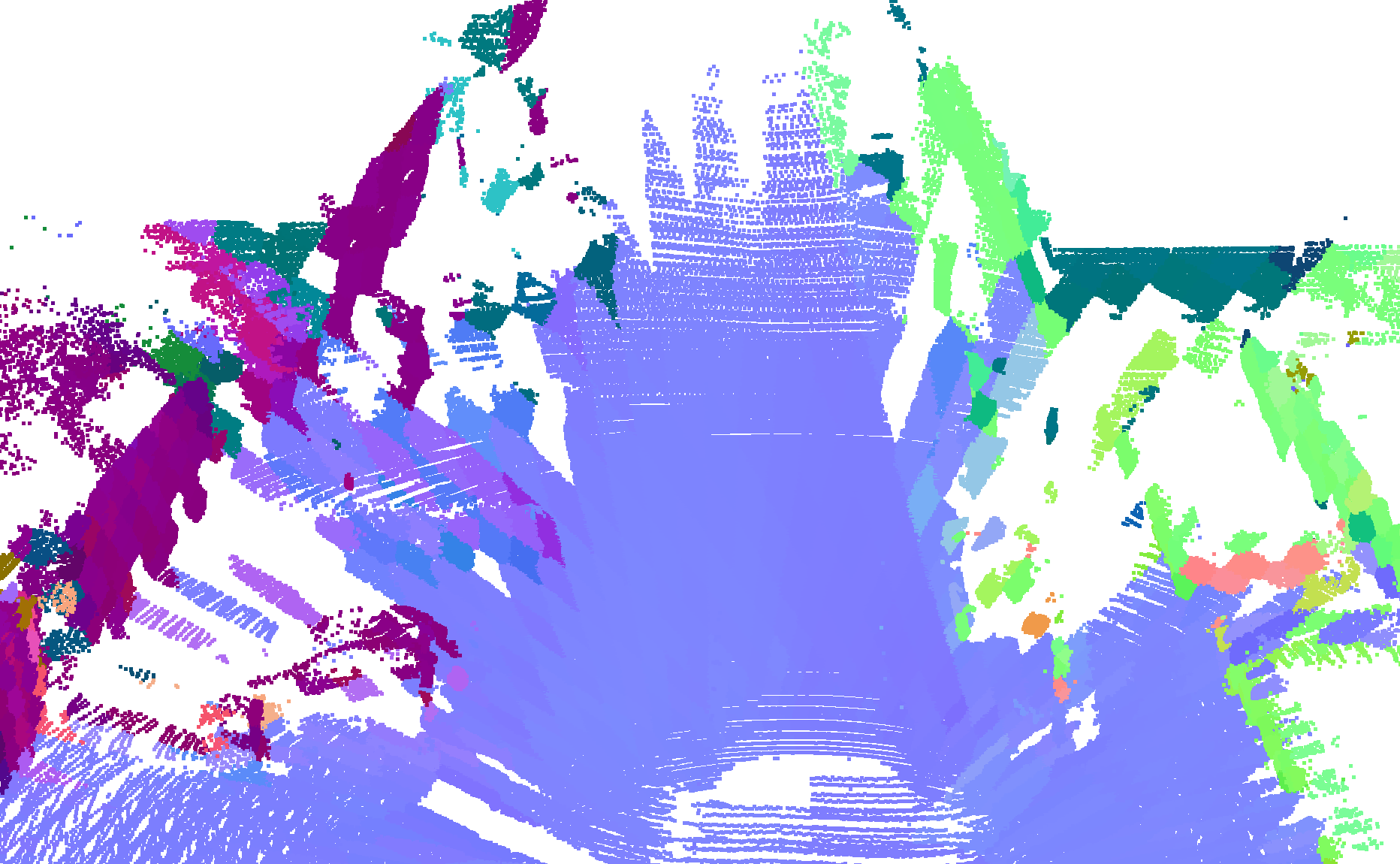}\label{fig:normal}}
    \end{minipage}  
    \vspace{-2em}
    \caption{An example of \core generation. Each local point cloud map is divided into segments according to $\mathcal{FL}$ shown in (b). The segmentation result is shown in (c), with different colors representing different segments. Then, plane extraction is performed for each segment to calculate its normal. We display the segments with valid planes in (d), colored according to their normal vectors.}  
    \vspace{-1.5em}
    
\end{figure} 
Inspired by TERRA~\cite{duan2024rotation}, a spherical descriptor for LiDAR calibration, we design a spherical structure to segment the map using a fixed method, which can recover the $\mathbb{R}^3$ position through one distance value. Specifically, the steps of \core generation from the local point cloud map are illustrated in the Fig.~\ref{fig:generation}, consisting of three steps as follows:

(1) First, a uniform spherical sampling algorithm Fibonacci lattice~\cite{Fibonacci} is used to generate a spherical point cloud on unit sphere denoted as $\mathcal{FL}\in\mathbb{R}^{N_{sp}\times3}$, which is unique in the whole algorithm. $N_{sp}$ represents the number of spherical samples, which is a hyperparameter. 
Additionally, we construct a KD-Tree $Tree_{\mathcal{FL}}$ for $\mathcal{FL}$, which is also the only KD-Tree used in the map construction process throughout the entire algorithm.

(2) Subsequently, the local map is transformed to the coordinate system of its initial frame, i.e., $\mathcal{M}'_{k_i} = \mathbf{P}^{-1}_{k_i} \cdot \mathcal{M}_{k_i}$, where we omit the augmentation. The moved $\mathcal{M}'_{k_i}$ is then normalized, i.e., projecting each point onto a unit sphere. Then, $Tree_{\mathcal{FL}}$ is utilized to find the closest point $\mathbf{p}^\mathcal{FL}$ in $\mathcal{FL}$ for each point in $\mathcal{M}'_{k_i}$. Points corresponding to the same $\mathbf{p}^\mathcal{FL}$ are grouped together for subsequent feature extraction. The grouped points are denoted as $\mathcal{G}_j$, where $j\in [1, N_{sp}]$, i.e.,  $\mathcal{M}'_{k_i}$ is segmented into $N_{sp}$ parts. An example of segmentation is shown in the Fig.~\ref{fig:segment}, where different colors represent different segments.

(3) For each $\mathcal{G}_j$, we use  RANSAC~\cite{schnabel2007efficient}  for plane extraction to calculate its normal. However, since the above segmentation method is based on angular approximation rather than the Euclidean distance, it results in many invalid planes formed by both foreground and background objects. Therefore, a simple clustering method is designed for this case. Specifically, we sort the group points by their distance to the origin, and if the distance difference between adjacent points exceeds a predefined threshold, 
the plane is deemed invalid. An example of plane extraction is shown in the Fig.~\ref{fig:normal} where the $\mathcal{G}_j$ with invalid plane is not shown.

(4) For $\mathcal{G}_j$ with valid plane, we calculate the distances $d_j$ between the origin and the intersection point where the plane and the ray formed by the origin and the $\mathcal{FL}$ point $\mathbf{p}^\mathcal{FL}_j$ meet. 
Finally, we form \core $\sycore_{k_i}\in\mathbb{R}^{{N_{sp}}\times4}$ by saving four numbers for each index, which include the distance and the plane's normal. 
 
Notably, since each segmented part is independent, the most computation-intensive step(3) can actually be processed in parallel.

\subsection{\core-based Bidirection Registration}
\label{sec：biregistration}
\begin{figure}[t]
	\centering
	\includegraphics[width = .9\linewidth]{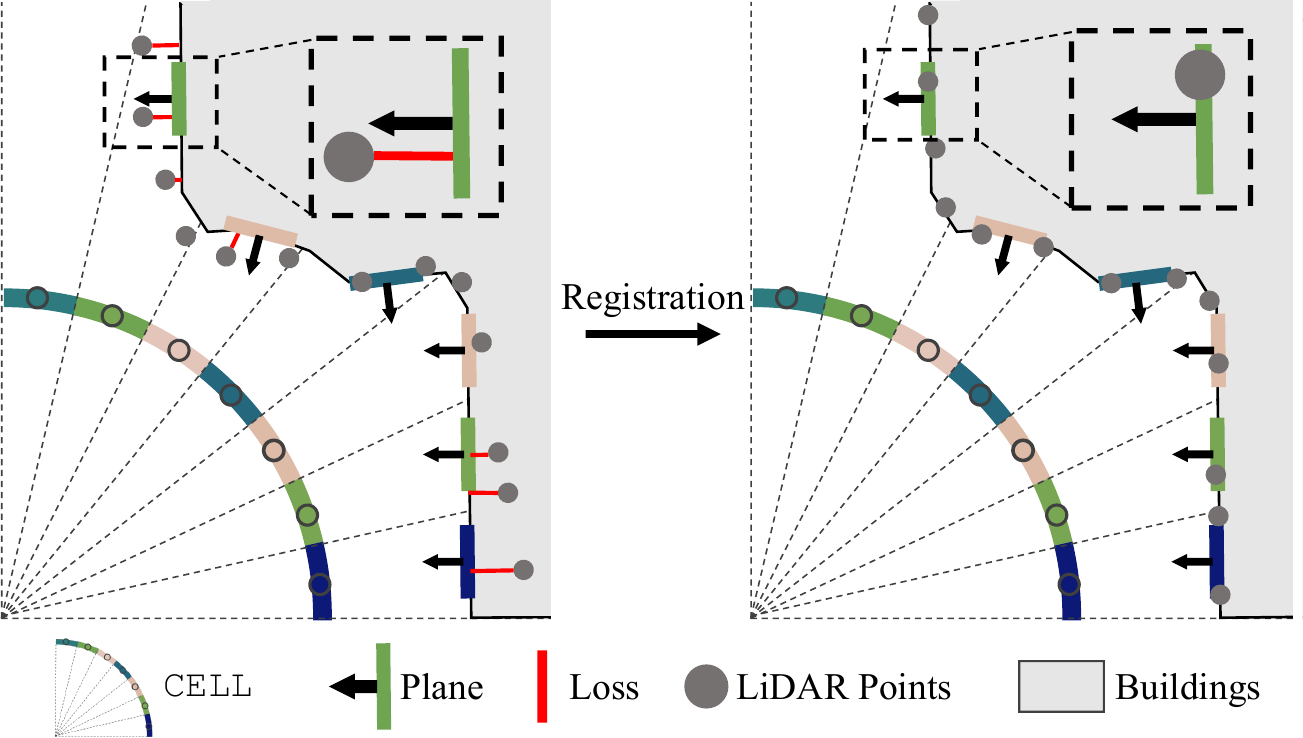}
	\caption{Scan-to-\core registration. First, the planes in the map are reconstructed according to the plane attributes and spherical point cloud $\mathcal{FL}$. Correspondences between LiDAR points and planes are established by finding the nearest points in $\mathcal{FL}$. Then, point-to-plane registration is performed to solve for the LiDAR frame's pose.
	}
    \vspace{-1.5em}
	\label{fig:registration}
\end{figure}

Based on the above description, \core contains geometry information of the local map and adjacent \cores can both contribute to estimating their relative poses intuitively. Based on this idea, we enhance the accuracy of the input LiDAR odometry by performing \core-based bidirectional registration using two adjacent \core and the two corresponding LiDAR scans. Before introducing bidirectional registration, we will first explain the scan-to-\core registration. Please note that our explanation is independent of the SLAM process and focuses solely on the registration algorithm itself. As shown in Fig.~\ref{fig:registration}, the registration takes an established \core and a LiDAR scan with a known initial pose $\mathbf{T}$ relative to the \core as input and outputs  the precise pose of the scan relative to the \core.

As described in Sec.~\ref{sec：generation}, \core stores the plane parameters fitted for each point cloud segment, so scan-to-\core registration is essentially a variant of point-to-plane registration. However, the main difference is that we reuse the $Tree_{\mathcal{FL}}$ to find correspondences on the sphere, rather than building a KD-Tree for each map to locate the nearest points in 3D space as correspondences.
Specifically, similar to step (2) in \core generation, for a point $\mathbf{p}^\mathcal{S}_i$ in the scan, it is first projected onto the unit sphere, and then the nearest point $\mathbf{p}^\mathcal{FL}_j$ in $\mathcal{FL}$ is found using $Tree_{\mathcal{FL}}$. The corresponding normal $\mathbf{n}_j$ and distance $d_j$ stored in \core represent the plane associated with $\mathbf{p}^\mathcal{S}_i$. The point-to-plane distance is calculated using the following formula:
\begin{equation}
\begin{aligned}
    D_{plane}(\mathbf{T}) = (\mathbf{T} \mathbf{p}^\mathcal{S}_i   - d_j \mathbf{p}^\mathcal{FL}_j )* \mathbf{n}_j
\end{aligned}
\end{equation}
The optimal pose estimation can be derived by solving the non-linear equation through Gauss-Newton method. The Jacobian matrix can
be solved by the chain rule as follows:
\begin{equation}
\begin{aligned}
    \mathbf{J}_d &= \frac{\partial\,  D_{plane}(\mathbf{T}\mathbf{p}^\mathcal{S}_i)}{\partial \mathbf{T}\mathbf{p}^\mathcal{S}_i}
    \frac{\partial\, \mathbf{T}\mathbf{p}^\mathcal{S}_i}{\partial \delta \xi}= \mathbf{n}_j \mathbf{J}_p,
\end{aligned}
\end{equation}
where $\mathbf{J}_p$ can be estimated by applying left perturbation model with $\delta \xi \in \mathfrak{ se}(3)$:
\begin{equation}
\label{eq:forward}
\begin{aligned}
    \mathbf{J}_p = \frac{\partial\, \mathbf{T}\mathbf{p}^\mathcal{S}_i}{\partial \delta \xi} &= \lim_{\delta \xi \rightarrow 0}\frac{exp(\delta \xi)\mathbf{T}\mathbf{p}^\mathcal{S}_i - \mathbf{T}\mathbf{p}^\mathcal{S}_i}{\delta \xi}\\
    &=\begin{bmatrix} \mathbf{I}_{3\times3}  & -[\mathbf{T}\mathbf{p}^\mathcal{S}_i]_\times \\ \mathbf{0}_{1\times3} & \mathbf{0}_{1\times3} \end{bmatrix},
\end{aligned}
\end{equation}
where $[\,]_\times$ means the skew symmetric matrix of the transformed point.

In practical use, we extend the scan-to-\core registration to bidirectional registration, leveraging two adjacent \cores and their corresponding frames. 
Specifically, to solve for the poses between $\sycore_{k_i}$ and $\sycore_{k_{i+1}}$, we perform both forward and reverse registration simultaneously. For forward registration, we use the pose $\mathbf{T} = \textbf{P}^{-1}_{k_i}\textbf{P}_{k_{i+1}}$ from the LiDAR odometry as the initial pose and employ scan-to-\core registration with $\mathcal{S}_{k_{i+1}}$ and $\sycore_{k_i}$. For reverse registration, we need to use $\textbf{T}^{-1}$ as the initial pose and perform  scan-to-\core registration with $\mathcal{S}_{k_i}$ and $\sycore_{k_{i+1}}$. To enable the reverse optimization process to run concurrently with the forward process, we rewrite the Eq.~\ref{eq:forward}. By applying a left perturbation to the inverse pose
\begin{equation}
\begin{aligned}
    \mathbf{T}'^{-1} = (\exp(\delta \xi^\land) \mathbf{T})^{-1} 
= \mathbf{T}^{-1}\exp(-\delta \xi^\land),\\
\end{aligned}
\end{equation}
the Jacobian matrix is calculated as follows:
\begin{equation}
\begin{aligned}
    \mathbf{J}_p = \frac{\partial\, \mathbf{T}^{-1}\mathbf{p}^\mathcal{S}_i}{\partial \delta \xi} &= \lim_{\delta \xi \rightarrow 0}\frac{(\mathbf{T}^{-1}(\mathbf{I}-\delta \xi^\land))\mathbf{p}^\mathcal{S}_i - \mathbf{T}^{-1}\mathbf{p}^\mathcal{S}_i}{\delta \xi}\\
    &=-\mathbf{R}^{-1}\begin{bmatrix} \mathbf{I}_{3\times3}  & -[\mathbf{p}^\mathcal{S}_i]_\times \\ \mathbf{0}_{1\times3} & \mathbf{0}_{1\times3} \end{bmatrix}.
\end{aligned}
\end{equation}

Since establishing correspondences and calculating loss can be parallelized, and there is no need to create a new KD-Tree for each registration, \core-based registration is highly efficient. This enables us to involve all points from both frames in the optimization process and iterate multiple times to guarantee the correctness of the correspondences and the convergence of the optimization, thereby ensuring the accuracy of \core-based registration.
 
\subsection{Loop Closure}
\label{sec：loop_detecton}
 
For loop detection, we implement a Euclidean distance-based approach similar to LIOSAM~\cite{shan2020lio}. Specifically, when a \core $\sycore_{k_i}$ is established, we search for the nearest $n_{loop}$ \cores within the current \coremap. Among these, \cores that do not belong to $k_{i-n_{loop}}$ to $k_i$ are considered loop candidates. Subsequently, scan-to-\core registration is performed between $\mathcal{S}_{k_i}$ and these candidates. During the registration process, those with an inlier ratio higher than a predefined threshold are deemed to be correct loop closures, the result will be added to the pose graph for global optimization.

\section{Experiments}
\begin{table}[t]
\center
\caption{Parameters and Their Values Used in Our Method}
\label{table:parameters}
\begin{tabular}{l|c}
\hline
Parameters Descripton                                     & value \\ \hline
The distance between two \cores, used in Sec.~\ref{sec：generation} & 6m     \\
$N_{sp}$, spherical sampling number  used in Sec.~\ref{sec：generation}                                & 50000 \\
Plane valid threshold used in Sec.~\ref{sec：generation}                        & 3     \\
Number of iterations in registration, used in Sec.~\ref{sec：biregistration}                                 & 4     \\
Loop candidates number, used in Sec.~\ref{sec：loop_detecton}                              & 10    \\
Inlier ratio threshold, used in Sec.~\ref{sec：loop_detecton}                         & 0.2  \\
\hline
\end{tabular}
\vspace{-1em}
\end{table}

\begin{table*}[t]
\center

\caption{ATE(\%) of LiDAR Odometry on KITTI Dataset. ``+'' means the trajectories are optimized by bidirectional registration.}
\label{table:kitti}
\resizebox{\linewidth}{!}{
\begin{tabular}{c|ccccccccccc|c|c}
\hline
Methods   & 0    & 1    & 2    & 3    & 4    & 5    & 6    & 7    & 8    & 9    & 10   & Mean &  $\downarrow(\%)$                    \\ \hline
A-LOAM    & 0.70 & 1.96 & 4.54 & 0.94 & 0.72 & 0.50 & 0.60 & 0.43 & 1.04 & 0.73 & 1.01 & 1.20 & \multirow{2}{*}{18.88\%} \\
A-LOAM+   & 0.60 & 1.33 & 4.29 & 0.64 & 0.53 & 0.37 & 0.38 & 0.36 & 0.90 & 0.53 & 0.74 & 0.97 &                          \\ \hdashline
F-LOAM~\cite{wang2021f}    & 0.70 & 1.95 & 1.04 & 0.95 & 0.69 & 0.49 & 0.54 & 0.42 & 0.93 & 0.70 & 1.03 & 0.86 & \multirow{2}{*}{26.88\%} \\
F-LOAM+   & 0.58 & 1.28 & 0.74 & 0.63 & 0.50 & 0.37 & 0.33 & 0.34 & 0.84 & 0.50 & 0.78 & 0.63 &                          \\ \hdashline
PFilter~\cite{duan2022pfilter}   & 0.90 & 0.68 & 0.80 & 0.82 & 0.42 & 0.66 & 0.27 & 0.51 & 1.08 & 0.64 & 0.76 & 0.69 & \multirow{2}{*}{12.52\%} \\
PFilter+  & 0.73 & \textbf{0.60} & 0.67 & 0.75 & 0.41 & 0.47 & 0.29 & 0.42 & 0.99 & 0.55 & 0.72 & 0.60 &                          \\ \hdashline
MULLS~\cite{pan2021mulls}     & 0.52 & 0.71 & 0.60 & \textbf{0.61} & 0.47 & 0.32 & 0.30 & 0.32 & \textbf{0.81} & 0.55 & 0.64 & 0.53 & \multirow{2}{*}{1.97\%}  \\
MULLS+    & 0.53 & 0.66 & 0.59 & 0.63 & 0.45 & 0.30 & 0.31 & \textbf{0.30} & 0.82 & 0.52 & 0.63 & 0.52 &                          \\ \hdashline
KISS-ICP~\cite{vizzo2023kiss}  & 0.52 & 0.72 & 0.53 & 0.66 & \textbf{0.36} & 0.30 & \textbf{0.26} & 0.33 & 0.82 & 0.49 & \textbf{0.56} & 0.50 & \multirow{2}{*}{2.96\%}  \\
KISS-ICP+ & \textbf{0.51} & \textbf{0.60} & \textbf{0.50} & 0.65 & 0.37 & \textbf{0.26} & 0.28 & \textbf{0.30} & 0.83 & \textbf{0.47} & 0.60 & \textbf{0.49} &                       \\ \hline
\end{tabular}}
\vspace{-1em}
\end{table*}

In the experiment, we first test the proposed bidirectional registration and loop detection algorithm on the KITTI dataset~\cite{kitti}, which is one of the most popular datasets for evaluating visual or LiDAR-based SLAM. Using five different LiDAR odometry methods as the baseline, we demonstrate the versatility of our backend. We then report the storage space required by \coremap compared to other common map representations. Finally, we present the time required for each step of our methods. The parameters and their values used in our
method are as shown in the Tab.~\ref{table:parameters}. All experiments are conducted on a laptop with an AMD Ryzen 7 4800H CPU and 16GB memory. 
\begin{table}[t]
\center
\caption{ATE(m) of LiDAR SLAM on KITTI Dataset. ``++'' means that the complete backend is used to optimize trajectories. }
\label{table:kitti_loop}
\begin{tabular}{c|ccccccc|c}
\hline
Methods    & 0   & 2   & 5   & 6   & 7   & 8   & 9   & Mean \\ \hline
MULLS~\cite{pan2021mulls}      & 1.1 & 5.4 & 1   & 0.3 & 0.4 & 2.9 & 2.1 & 1.9  \\
SuMa~\cite{suma}
& 1   & 7.1 & 0.6 & 0.6 & 1   & 3.4 & 1.1 & 2.1  \\
Litamin2~\cite{yokozuka2021litamin2}   & 1.3 & 3.2 & 0.6 & 0.8 & 0.5 & 2.1 & 2.1 & 1.5  \\
HBA~\cite{liu2023large}        & 0.8 & 5.1 & 0.4 & 0.2 & 0.3 & 2.7 & 1.3 & 1.5  \\
PIN-SLAM~\cite{pan2024pin}   & 0.8 & 3.3 & 0.2 & 0.4 & 0.3 & 1.7 & 1   & 1.1  \\ \hdashline
KISS-ICP+ & 4.4 & 7.5 
 & 1.6 
 & 0.5 
 & 0.3 & 2.4 
 & 1.5 
 & 2.6 \\
KISS-ICP++ & 0.8 & 4.8 & 0.3 & 0.3 & 0.3 & 1.8 & 1.3 & 1.4  \\
\hline
\end{tabular}
\end{table}
\begin{table}[t]
\center
\caption{Map Memory(MB). ds represents downsample.}
\label{table:map_memory}
\begin{tabular}{cccc}
Representation         & KITTI 00 & KITTI 05 & KITTI 08 \\ \hline
Raw Point Cloud            & 8446.1  &  5284.7 & 7640.2 \\
Point Cloud(ds 0.4m)   & 109.6 &  53.1 &  172.1 \\
Surfel (SUMA)       & 887.7   &  512.6    & 835.7    \\
Mesh (PUMA)         &2032.9     & 1317.4    & 1894.1  \\
Implicit Neural (PIN-SLAM)  & 102.1   &  66.3   & 138.8  \\ \hdashline
\coremap                &  \textbf{51.2+9.0} 
 & \textbf{35.3+6.6} 
  & \textbf{37.5+6.7} 
   \\ \hline

\end{tabular}
\vspace{-2em}

\end{table}

\subsection{Trajectory Accuracy Evaluation}
\subsubsection{LiDAR Odometry Evaluation on KITTI}
 
First, we test the \core-based bidirectional registration. As shown in Tab.~\ref{table:kitti}, we evaluate five commonly used LiDAR odometry methods with varying performance on the 11 sequences (00-10) of the KITTI dataset. Following \cite{zheng2024traj}, a vertical angle of $0.205^\circ$ is used to rectify the calibration errors in raw point clouds. We use average translational error, ATE (m/100m), as in ~\cite{kitti} to evaluate the accuracy of LiDAR odometry. 
The baseline data for the five methods are obtained by running the latest versions of their respective codes from GitHub. The data ending with a ``+'' are obtained by running the corresponding algorithms with \core-based bidirectional registration. As indicated in the mean error column of the table, our algorithm achieves positive improvements for all methods. For the originally less effective A-LOAM, F-LOAM~\cite{floam}, and PFilter~\cite{duan2022pfilter}, the improvements exceeded 10\%. Meanwhile, for the already well-performing MULLS~\cite{pan2021mulls} and KISS-ICP~\cite{vizzo2023kiss}, there are still slight improvements. 
\subsubsection{LiDAR SLAM Evaluation on KITTI}
 
Next, we evaluate the performance of our loop detection algorithm on the KITTI dataset. Since the KITTI evaluation metric ATE (\%) only shows errors up to 800m, it cannot reflect the global consistency of the map. Therefore, we follow the testing method~\cite{pan2024pin} commonly used in LiDAR SLAM with loop detection, adopting the root mean square error (RMSE) of the absolute trajectory error (ATE) with Umeyama trajectory alignment as the localization accuracy metric. Tab.~\ref{table:kitti_loop} presents the accuracy of the KITTI sequences containing loops. In addition to showcasing our results, we also present the accuracy of some state-of-the-art methods, with data obtained from PIN-SLAM~\cite{pan2024pin}. 
As shown in the table, compared to the trajectory without loop, our loop detection method (ended by ``++'')  enhances the global consistency. Furthermore, KISS-ICP++ ranks second in accuracy, only behind PIN-SLAM.

\subsection{\core Memory}
\begin{table}[t]
\center
\caption{Runtime Analysis}
\label{table:runtime}
\begin{tabular}{ccc}
\hline
Module                                           & Submodule               & t(ms) \\ \hline
\multirow{2}{*}{\core Generation} & Segmentation            &  11.0   \\
                                                 & Plane extraction        &  49.0  \\ \hdashline
\multirow{2}{*}{Bidirectional Registration}      & Finding correspondences &  384.9      \\
                                                 & Pose estimation         &  676.6     \\ \hdashline
\multirow{3}{*}{Loop Closure}                    & Finding 
 candidates              &   9.7    \\
 & Finding correspondences &  183,3     \\
                                                 & Pose estimation         &  326.2  \\ \hline

\end{tabular}
\vspace{-2em}

\end{table}
 
Using three longer sequences from the KITTI dataset as examples, we demonstrate the lightweight nature of \coremap. As mentioned in Sec.~\ref{sec：generation}, we store the parameters of $N_{sp}$ planes in each \core, along with the corresponding poses. However, in practice, most of the $N_{sp}$ regions are either not scanned by the LiDAR or with invalid planes, making it meaningless to store their parameters. Therefore, we further compress the storage space by additionally storing the index of each valid plane region. As shown in Tab.~\ref{table:map_memory}, the storage of \coremap is displayed in two parts: the first half for plane parameters and the second half for the index list corresponding to each plane. The pose of each \core is also stored, but since it occupies relatively little space, it is omitted from the table.
We present the space occupied by some common map representations including point cloud, surfel~\cite{suma}, mesh~\cite{vizzo2021icra}, and implicit neural~\cite{pan2024pin}. Since \core does not store the specific locations of the scene but can be reconstructed through unique spherical point clouds $\mathcal{FL}$ and plane parameters, \coremap uses the least amount of space to store the global map's geometric structure.
\subsection{Runtime Analysis}
As shown in Tab.~\ref{table:runtime}, using KITTI 00 as an example, we demonstrate the time required for each module in the complete algorithm.
The \core generation module and bidirectional registration module are executed once each time a \core is established. In the bidirectional registration module, all point clouds from the two LiDAR frames participate in the registration, and the steps of finding correspondences and pose estimation are iterated four times, taking approximately 1s in total. Since the \core establishment interval is set to 6m, when the LiDAR moves at a speed below 6m/s, the bidirectional registration steps will not cause cumulative delay; however, if the speed exceeds this, it will. This can be mitigated by appropriately downsampling the point clouds or reducing the number of iterations. For loop detection, since only scan-to-\core registration is required, the time taken is about half that of the bidirectional registration module. Additionally, pose graph optimization is executed in another thread and runs in parallel with the modules listed in the table, so its time consumption is not included.
\section{Conclusion}
We present \coremap, an elastic and lightweight map representation, and develop a general backend for LiDAR SLAM based on the proposed representation. Our approach improves LiDAR SLAM performance and global map consistency by implementing a bidirectional registration module and a loop closure detection module. The experiments demonstrate that \coremap substantially enhances the existing solutions regarding trajectory accuracy and map memory. Moving forward, we will explore the application of \coremap in downstream tasks such as navigation. Additionally, we will leverage \core's lightweight characteristics to explore its use in large-scale collaborative mapping with multiple robots.

\bibliographystyle{IEEEtran}
\bibliography{IEEEabrv, references}

\end{document}